\documentclass{article}
\usepackage{stywhispers,amsmath,epsfig}
\usepackage{tocloft}
\usepackage{hyperref}
\usepackage{setspace}
\usepackage{float}
\usepackage{graphicx}
\usepackage{subfig}
\usepackage{lastpage}
\usepackage{bbold}
\usepackage{placeins}
\usepackage{enumitem}
\usepackage{listings}
\usepackage{multirow} 
\usepackage{array}
\usepackage{lscape} 
\usepackage{longtable} 
\usepackage[T1]{fontenc}
\usepackage{fancyhdr,lipsum}
\usepackage{cite}
\usepackage{amsmath,amssymb,amsfonts}
\usepackage{algorithmic}
\usepackage{graphicx}
\usepackage{textcomp}
\usepackage{xcolor}
\usepackage{subcaption}
\usepackage[export]{adjustbox}

\title{Manifold Learning for Hyperspectral Images\thanks{{ Author Email: \small{fethi.harkat@univ-grenoble-alpes.fr}, 
}
\small Code available at\cite{code}
 }}

\name{Fethi Harkat\textsuperscript{\large*} \textsuperscript{\large†},   Guillaume Gey\textsuperscript{\large*}, Val\'erie Perrier\textsuperscript{\large†}, Kévin Polisano\textsuperscript{\large†},  Tiphaine Debeuret\textsuperscript{\large*},}

\address{\textsuperscript{\large*}Detection Technology Plc, A Grid, Otakaari 5A, 02150 Espoo, Finland \\
\textsuperscript{\large†} Univ. Grenoble Alpes, CNRS, Grenoble INP, LJK, 38000 Grenoble, France}

\begin{document}


\maketitle

\begin{abstract}
    Traditional feature extraction and projection techniques, such as Principal Component Analysis, struggle to adequately represent X-Ray Transmission (XRT) Multi-Energy (ME) images, limiting the performance of neural networks in decision-making processes. To address this issue, we propose a method that approximates the dataset topology by constructing adjacency graphs using the Uniform Manifold Approximation and Projection. This approach captures nonlinear correlations within the data, significantly improving the performance of machine learning algorithms, particularly in processing Hyperspectral Images (HSI) from X-ray transmission spectroscopy. This technique not only preserves the global structure of the data but also enhances feature separability, leading to more accurate and robust classification results.
   
\end{abstract}



\section{Introduction}
Recent advances in Hyperspectral Images (HSI) analysis have primarily focused on reflection spectroscopy in the visible or near-infrared light domains. However, these methods often underperform in X-ray transmission (XRT) spectroscopy \cite{ou2021recent} due to the distinct noise characteristics in this energy domain, the low-energy resolution of X-ray detectors, and poor counting statistics. Traditional methods, such as the Principal Component Analysis (PCA) \cite{pearson1901}, and Non-negative Matrix Factorization (NMF) \cite{lee2000algorithms} have shown limitations in characterizing these datasets, which in turn affects the efficiency of neural networks. Over the past decade, manifold learning approaches have gained increasing interest for HSI analysis \cite{6678226}. These approaches typically aim to project data onto a low-dimensional space to reveal intrinsic structures, reduce noise, and improve classification performance. Studies have demonstrated the effectiveness of manifold-based techniques, such as t-distributed Stochastic Neighbor Embedding (t-SNE) \cite{vanDerMaaten2008} combined with Convolutional Neural Networks (CNN) \cite{lecun1995convolutional}, in enhancing non-linear dimensionality reduction \cite{sorzano2014survey}\cite{jia2022feature} and classification performance \cite{1396318, 8876672}. In this work, we propose to apply the Uniform Manifold Approximation and Projection (UMAP) \cite{mcinnes_umap_2020} \cite{healy2024uniform} technique to X-ray HSI due to its ability to more accurately approximate the dataset’s topology, thus better capturing nonlinear correlations and improving the performance of subsequent machine learning algorithms.

\begin{figure}[htb]
    \centering
    \includegraphics[width=0.5\textwidth]{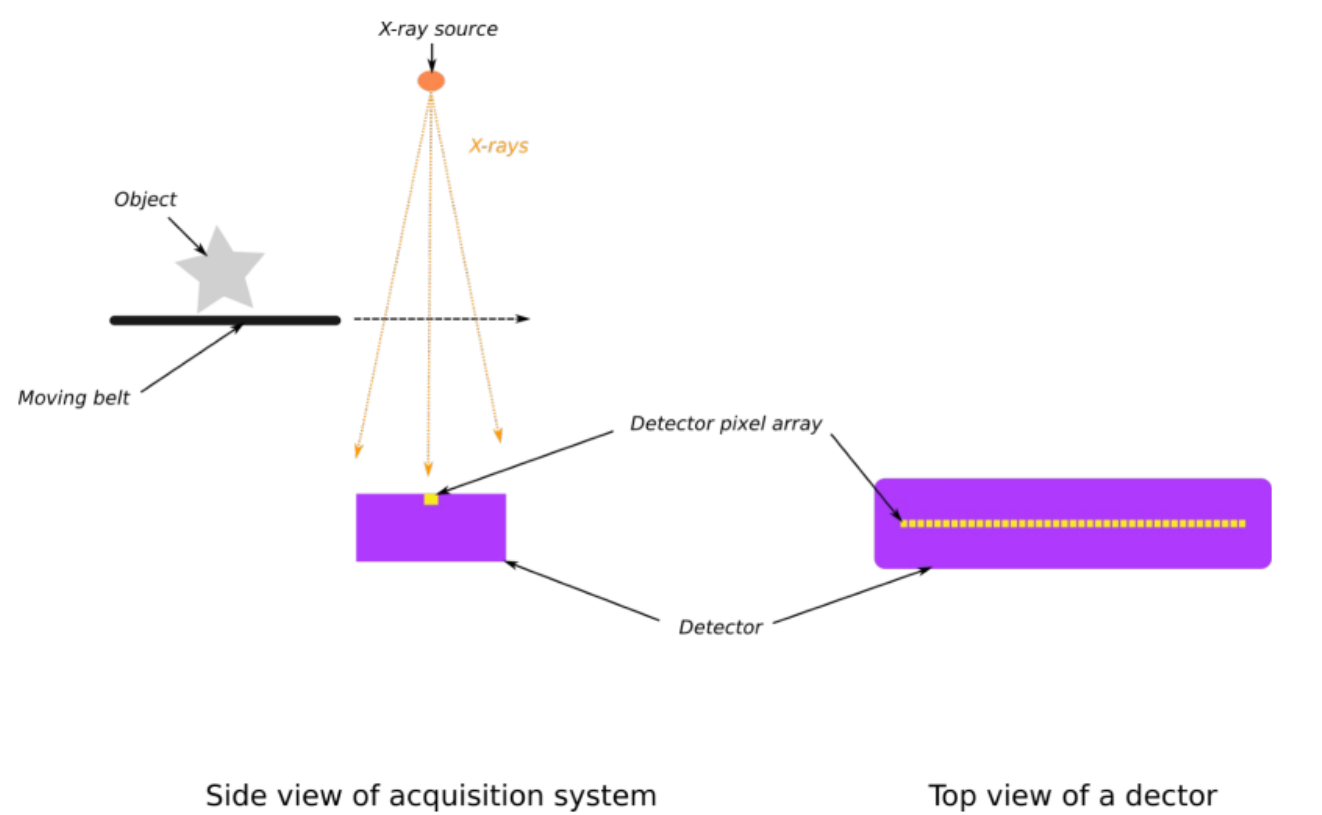}
    \caption{Scheme of the experimental setting (left) view from aside. A top view of the detector is 
also given (right) to make the data collection process clearer to the reader.}
    \label{fig}
\end{figure}

Our detector consists of an array of semi-conductor active areas, which we will refer to as “detector pixels” to avoid confusion with image processing terminology. Each detector typically contains 512 detector pixels. The data acquisition process involves using an X-ray source to illuminate an object placed on a moving belt, while a detector captures the X-rays transmitted through the object during its motion (Fig.~\ref{fig}). During the process, both the source and the detector remain static. The resulting images have a shape of $[H, W, C]$ where $H$ represents the time axis, $W$ the detector pixel axis, and $C$ the energy band axis.

This work focuses on three main objectives:
\begin{itemize}
\item Constructing an expressive representation of spectral data,
\item Reducing the dimensionality of this representation,
\item Evaluating it using deep learning, while also gaining insights into why this method is effective.
\end{itemize}
 
In this paper, we investigate the application of UMAP to XRT HSI data. The proposed methodology details the approaches and tools used in our analysis. In Section~\ref{section_UMAP}, we present the UMAP method and its application in our approach. Section~\ref{section_EXP} provides an overview of our experiments, and finally, in Section~\ref{section_Results}, we present our results and insights.

\section{Methodology}
\label{section_UMAP}
\subsection{UMAP:Uniform Manifold Approximation and Projection}
UMAP~\cite{mcinnes_umap_2020} is a powerful technique for dimensionality reduction and visualization, widely used in data analysis and visualization \cite{becht2019dimensionality,diaz2021review}. It serves as an alternative to t-SNE, with the added advantage of preserving both local and global structures in high-dimensional data. UMAP has gained popularity due to its ability to retain meaningful relationships between data points while remaining computationally efficient.

The method is based on manifold learning, which assumes that data lies on a lower-dimensional manifold embedded on a high-dimensional space. UMAP seeks to find a lower-dimensional representation that preserves the intrinsic structure of data. This is achieved by constructing a fuzzy topological representation of neighborhood relationships. Using fuzzy set theory \cite{zimmermann2011fuzzy}, UMAP estimates the probability that data points are neighbors. The algorithm then minimizes the cross-entropy between pairwise probabilities in the high-dimensional space and their counterparts in the low-dimensional embedding, ensuring that the global structure is effectively preserved.

In our experiments, we used the Parametric UMAP \cite{sainburg2021parametric} which employs a Neural Network (NN) to learn a mapping from the original data to its embedding while optimizing the same objective as traditional UMAP. This approach enables fast embedding of new data, as the trained network directly maps inputs without requiring re-optimization, making it particularly suitable for large or dynamic datasets.

\subsection{Our approach}
\begin{figure}[htb]
    \centering
    \includegraphics[scale=.80]{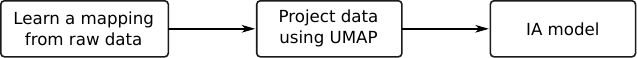}
    \caption{Pipeline of our proposed method.}
    \label{plan}
\end{figure}

Our approach (Fig. \ref{plan}) consists of the following steps:
\begin{enumerate}
    \item Fit the UMAP model with images of size $[H,W,C]$ where $H$, $W$ and $C$ are respectively the width, height and number of bands of raw images, with the target dimension $D$, where $D \leq C$.
    \item Project all images using the UMAP model onto the learnt lower-dimensional space of size $D$, and reshape them to their new shape $[H,W,D]$.
    \item Feed the projected images into a specific CNN model to perform various machine learning tasks such as segmentation or regression.
\end{enumerate}

\section{Numerical experiments}
\label{section_EXP}
To evaluate the effectiveness of our proposed methodology, we assess its performance across three different datasets and tasks. Through these experiments, we aim to demonstrate the adaptability and efficiency of our approach for both segmentation and regression challenges in hyperspectral imaging (HSI). Below, we describe each dataset's characteristics and the respective machine learning task.

\subsection{Segmentation of Cigarettes}
\label{exp_cigarettes}
The Cigarettes dataset was created internally and consists of HSI images of random luggage, which may contain either a pack or a carton of cigarettes (positive samples) or none (negative samples). The objective of this experiment is to segment cigarettes in the images when they are present. Binary masks of cigarettes inside the luggage have been carefully annotated for each relevant image.

\begin{figure}[htb]
        \centering
        \includegraphics[scale=.3]{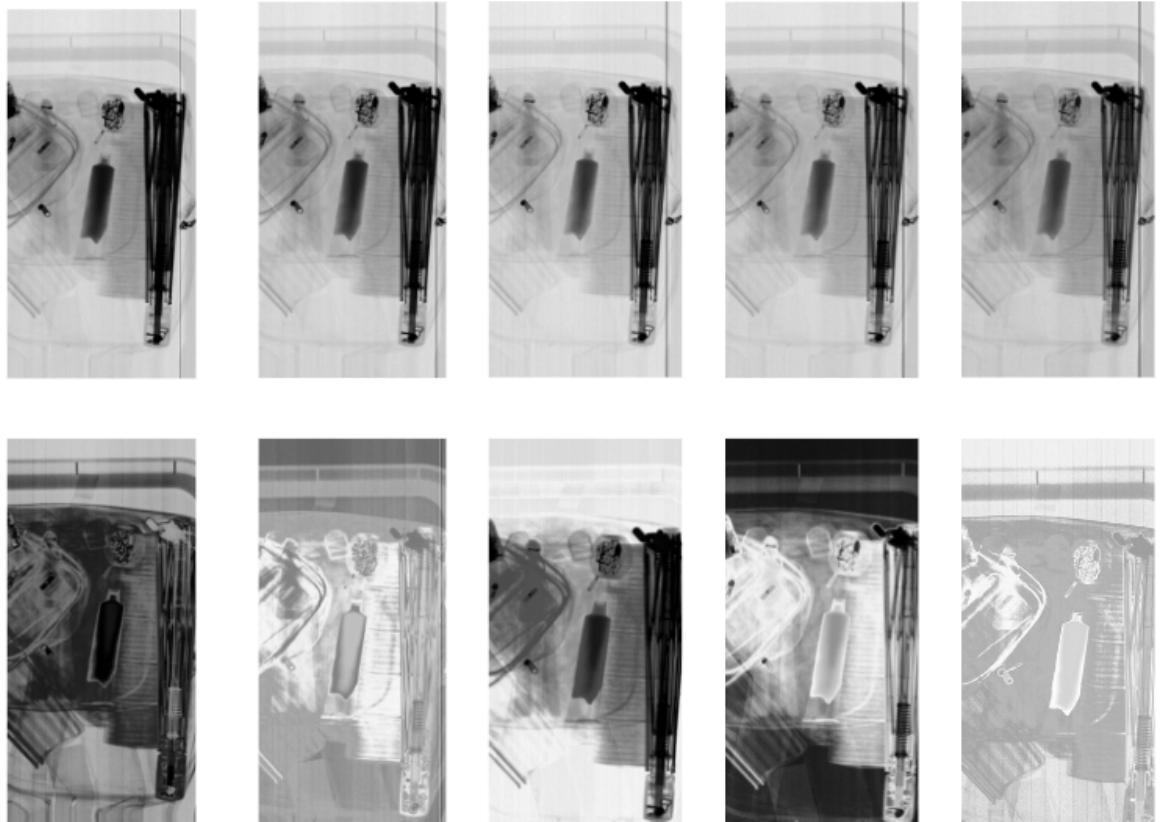}
        \caption{Comparison between raw spectral bands (top row) and UMAP-projected bands (bottom row).}
        \label{Raw}
\end{figure}

\begin{figure}[htb]
    \centering
    \resizebox{0.7\linewidth}{!}{
    \begin{tabular}{ >{\centering\arraybackslash}m{0.22\textwidth} 
                     >{\centering\arraybackslash}m{0.05\textwidth}
                     >{\centering\arraybackslash}m{0.15\textwidth} 
                     >{\centering\arraybackslash}m{0.05\textwidth}
                     >{\centering\arraybackslash}m{0.22\textwidth} }
        \includegraphics[width=\linewidth,valign=m]{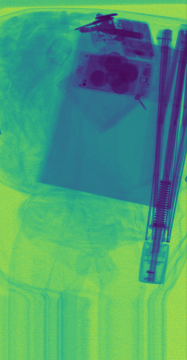} &
        $\times$ &
        \includegraphics[width=\linewidth,valign=m]{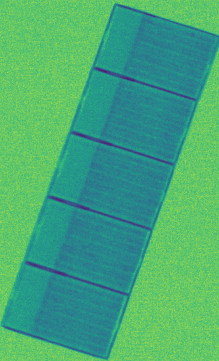} &
        $=$ &
        \includegraphics[width=\linewidth,valign=m]{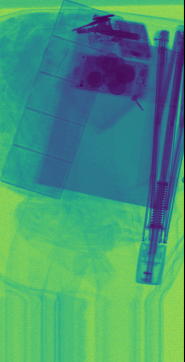}
    \end{tabular}
    }
    \caption{Illustration of the fusion process: normalized luggage and cigarette images are combined to generate synthetic data.}
    \label{fig:fusion}
\end{figure}

We aimed to expand our training set without the need for complex and costly image acquisition procedures. 
Following the Beer-Lambert law~\cite{miller2009beer}, which models photon attenuation in a medium, 
we generated synthetic training images by separately capturing luggage and cigarette images, applying white normalization, and combining them. 
Using this approach, we built a dataset of 320 training images and 30 synthetic test images see Fig \ref{fig:fusion}. 
In addition, 21 real test images were acquired with manually placed cigarette cartons inside luggage to evaluate the performance of our pipeline on real data. 
Results for both test sets are presented in Section~\ref{section_Results}.

To reduce the computational cost, the input images were cropped resized from $[16, 1000, 512]$ to $[16, 152, 128]$ using 2D average pooling. The trained Parametric UMAP then  projected the downsized images onto a subspace of dimension $D=5$ before feeding them into the CNN. A sample of these bands is shown in Figure~\ref{Raw}. 

\begin{figure}[htb]
    \centering
    \includegraphics[width=0.7\linewidth]{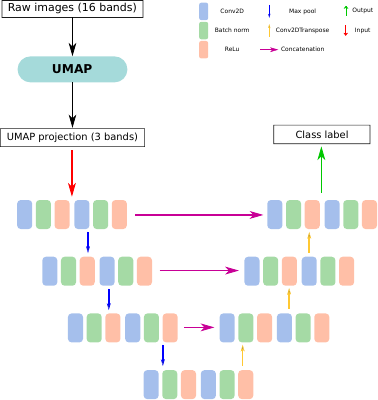}
    \caption{The U-Net architecture consists of four downsampling blocks, each with a Double Convolution  followed by Max Pooling  (stride = 2), progressively increasing the number of filters from 8 to 64. The upsampling blocks include an Upsampling (scale = 2), concatenation with the corresponding output, and a Double Convolution to refine features. The final convolution reduces the output to a single prediction channel.}
    \label{unet}
\end{figure}

For the segmentation task, a U-Net model \cite{azad2024medical} is used with Binary Cross Entropy as the loss function. The symmetric architecture (see Fig. \ref{unet}) of U-Net includes an encoder, capturing context by compressing the input into feature maps, and a decoder, reconstructing the spatial dimensions to produce the segmented output. UMAP hyperparameters where left to their default values. 

\subsection{Chemical composition of Stones}
\label{exp_stones}
\begin{figure}[htb]
    \centering
    \includegraphics[width=0.2\textwidth]{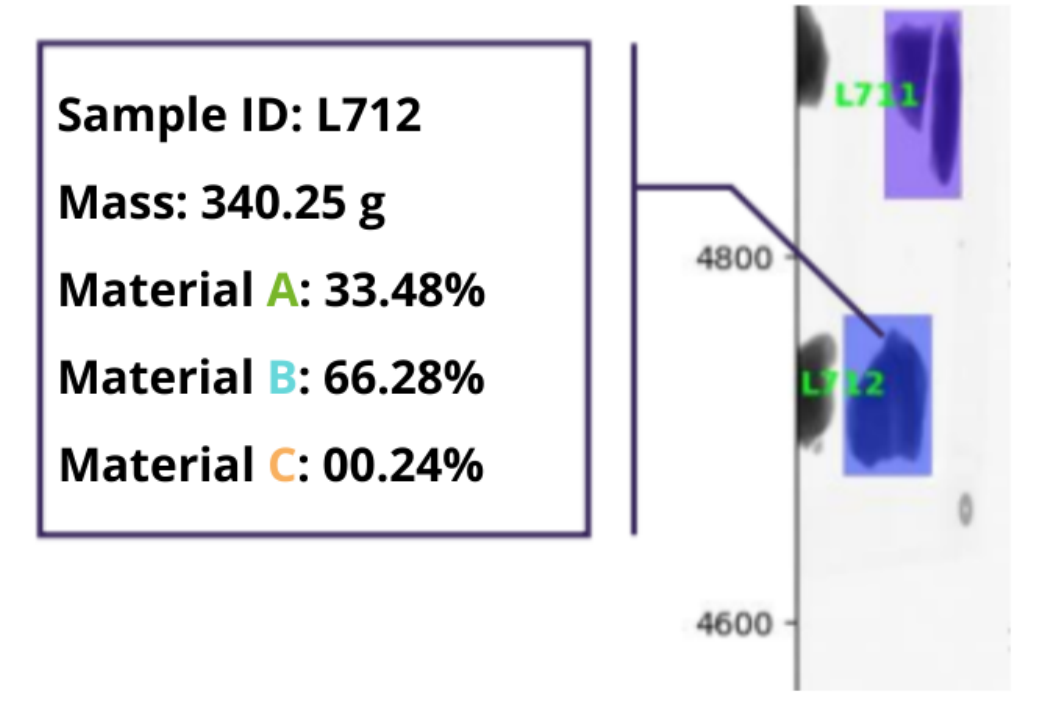}\label{labels}
    \caption{Labels of Stones: each stone has an ID along with information about its total mass and material concentrations. The material $C$ is residual elements and always has a very low concentration.} 
    \label{pil data}
\end{figure}

The Stones dataset consists of individual stone images from an industrial mine, each composed of two main chemical components. Each image includes a binary mask, mass (grams), and concentration percentages of components $A$ and $B$, with a third component $C$ accounting for residual elements (see Fig. \ref{pil data}). The goal is to predict mass and concentrations of $A$ and $B$ from the HSI and binary mask. $C$ can be derived from $A$ and $B$ and always has the lowest concentration, therefore it is not a relevant prediction target. As $B$ concentration dominates, standard error averaging is unsuitable for performance evaluation, necessitating a more balanced metric. To ensure a better estimation of performances over all the targets, we therefore define the \textit{Special Harmonic} (SH) score:
\begin{equation}
\label{eq:SH}
    \text{SH-score} = M ~\frac{\prod_{i=1}^{M} S_i}{\sum_{i=1}^{M} S_i}
\end{equation}
where $M$ is the number of regression targets, and $S_i$ the associated individual score. The \text{SH-score} as well as the $S_i$ are bounded to $[0,1]$, where 1 is a perfect result. $S_i$ are defined as follows:
\begin{equation}
\label{eq:S}
    S_i = \exp \left( -\frac{\sum_{j=1}^{N} |p_{ij} - t_{ij}|}{\sum_{j=1}^{N} t_{ij}} \right), ~
\end{equation}
where $p_{ij}$ and $t_{ij}$ respectively stands for the prediction and the target value associated to $i\textit{-th}$ regression target for $j\textit{-th}$ sample, $N$ being the total number of samples. The rationale behind this formulation of individual score $S_i$ is to collapse quickly in case of poor performances, thus giving a solid estimation of performance robustness through the SH-score. We decided to name this score as such because of its close resemblance with the \textit{Harmonic Mean} stripped from its cross-products in the denominator. To avoid any unnecessary complexity in this work, we decided to use a simple sum of weighted squared errors as the minimization objective:
\begin{equation}
    Loss = \sum_{i=1}^{M} w_i \left( \frac{p_{i} - t_{i}}{t_{i}} \right)^2
\end{equation}
where $p_{i}$, $t_{i}$ and $w_i$ respectively stands for the prediction, target and weight values associated to $i\textit{-th}$ regression target. The $w_i$ have been empirically set to $\left\{0.5, 5, 5 \right\}$, corresponding respectively to the mass, the material $A$ and $B$ targets.

The Stones dataset consists of 1,453 training images and 514 test images. To reduce computational load, 4\% of the dataset is used for Parametric UMAP training, selecting a balanced subset based on mass and concentrations. The UMAP hyperparameters are set to a minimum distance of 0.5 and 50 neighbors, the configuration found to work best empirically. The minimum distance controls point spacing, while the number of neighbors balances local and global structures.

\begin{figure}[htb]
    \centering
    \includegraphics[width=0.2\textwidth]{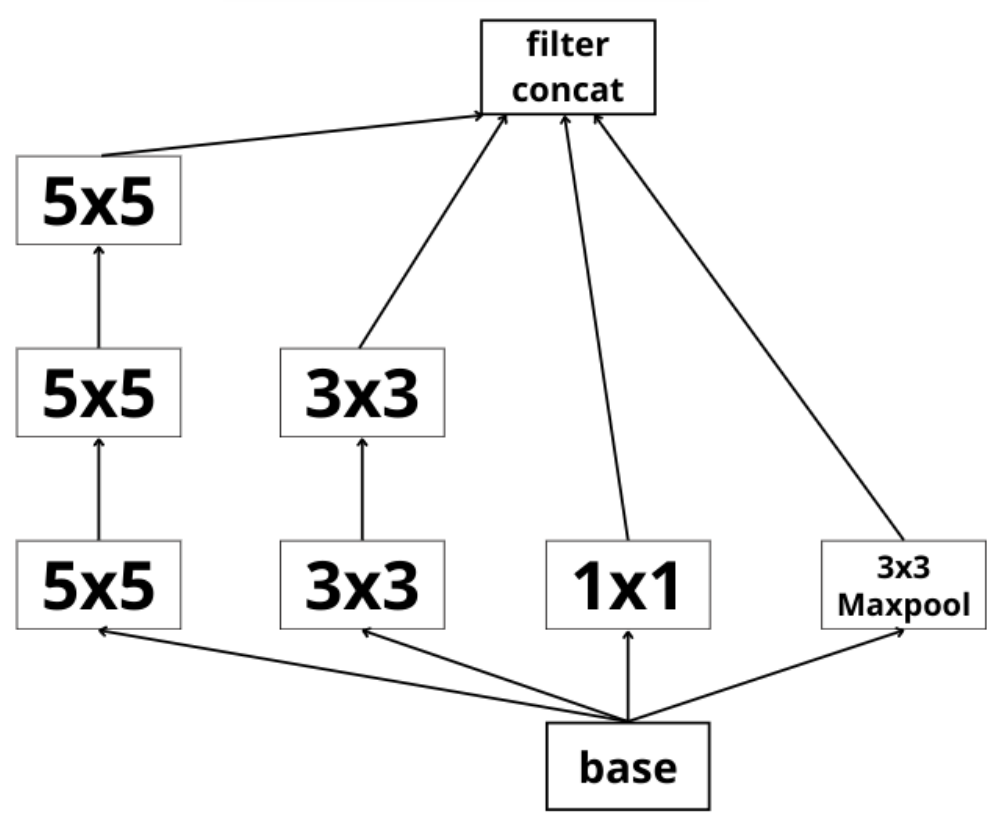}\label{inception}
    \caption{Customized filters used for the regression task.} 
    \label{incep}
\end{figure}

We tested UMAP projections with 10, 20, and 30 bands, feeding the results into a CNN encoder of four Inception \cite{szegedy2016rethinking} blocks (Figure~\ref{incep}) followed by a task-specific decision layer. All 514 test images were projected using the fitted UMAP models, and results were compared to the CNN trained on 64 raw bands, 20-band PCA, and 20-band NMF.

\subsection{Indian Pines dataset}
The Indian Pines dataset \cite{IndianPines}, captured by the AVIRIS sensor in 1992 over Indiana, consists of a 145$\times$145 pixel image with 220 spectral bands. It includes 16 land-cover classes, with labeled ground truth for part of the image, and is widely used to evaluate hyperspectral classification and feature extraction methods due to its high dimensionality and complexity.

\begin{figure}[htb]
    \centering
    \includegraphics[width=0.2\textwidth]{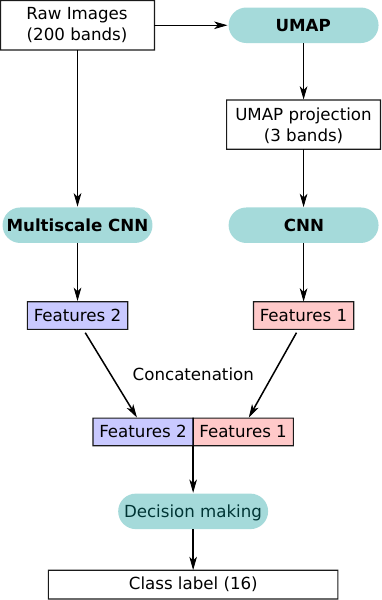}
    \caption{Multiscale CNN model with two parallel branches and their respective inputs, the UMAP projection and the raw image.}
    \label{class}
\end{figure}

For this experiment, we used a  multiscale CNN model \cite{li2019multi} with two branches and two inputs (see Fig. \ref{class}). Multiscale feature fusion help leveraging diverse spatial structures and rich texture features through extensive neighborhood associations. As before, data were projected onto a low dimension representation using UMAP, then fed into the CNN architecture. Results were compared to those obtained using t-SNE and PCA as data reduction techniques. In all cases, the target dimension was set to $D = 3$. UMAP hyperparameters where left to their default values for this experiment as well. 

\section{Results}
\label{section_Results}
\subsection{Cigarettes dataset}
As mentioned in Section~\ref{exp_cigarettes}, we evaluated our approach on both real and synthetic datasets using IoU and Dice Score as segmentation metrics \cite{muller2022towards}. Each method was run 10 times to compute error bars, and the averaged results are summarized in Table~\ref{tab:combined_results}.

\begin{table}[htbp]
\centering
\small
\resizebox{\columnwidth}{!}{%
\begin{tabular}{|l|l|c|c|c|c|c|}
\hline
\textbf{Dataset} & \textbf{Method} & \textbf{Bands} & \textbf{Mean IoU} & \textbf{IoU Err} & \textbf{Mean Dice} & \textbf{Dice Err} \\ 
\hline
\multirow{5}{*}{Synthetic} 
  & Raw Data   & 16 & 0.766 & 0.032 & 0.865 & 0.005 \\ 
  & NMF        & 5  & 0.214 & 0.012 & 0.330 & 0.009 \\ 
  & PCA     & 5  & 0.417 & 0.012 & 0.572 & 0.005 \\ 
  & PCA     & 8  & 0.433 & 0.010 & 0.598 & 0.007 \\ 
  & \textbf{UMAP} & \textbf{5} & \textbf{0.833} & \textbf{0.016} & \textbf{0.901} & \textbf{0.006} \\ 
\hline
\multirow{4}{*}{Real} 
  & Raw Data   & 16 & 0.230 & 0.020 & 0.370 & 0.010 \\ 
  & NMF        & 5  & 0.200 & 0.030 & 0.330 & 0.010 \\ 
  & PCA        & 5  & 0.180 & 0.030 & 0.300 & 0.010 \\ 
  & \textbf{UMAP} & \textbf{5} & \textbf{0.320} & \textbf{0.040} & \textbf{0.490} & \textbf{0.010} \\ 
\hline
\end{tabular}%
}
\caption{Comparison of segmentation performance across synthetic and real datasets for different dimensionality reduction methods.}
\label{tab:combined_results}
\end{table}


Table~\ref{tab:combined_results} shows that our UMAP-based approach clearly outperforms PCA, NMF, and even the raw data, demonstrating its ability to capture non-linear spectral structures while removing noise and redundancy. PCA slightly improves over NMF, but both linear methods fall far behind UMAP, confirming that linear projections are insufficient for complex hyperspectral XRT data. Since all methods were evaluated using the same number of bands, the comparison is fair. 

UMAP significantly outperforms PCA and NMF on both datasets, even with fewer bands. The lower scores on the real dataset reflect the crude approximations used to compensate for limited data. Despite this gap, the ranking of methods remains consistent, confirming the robustness of UMAP.

\begin{figure}[htbp]
  \centering
  \begin{minipage}[t]{0.48\columnwidth}
    \centering
    \includegraphics[width=\linewidth]{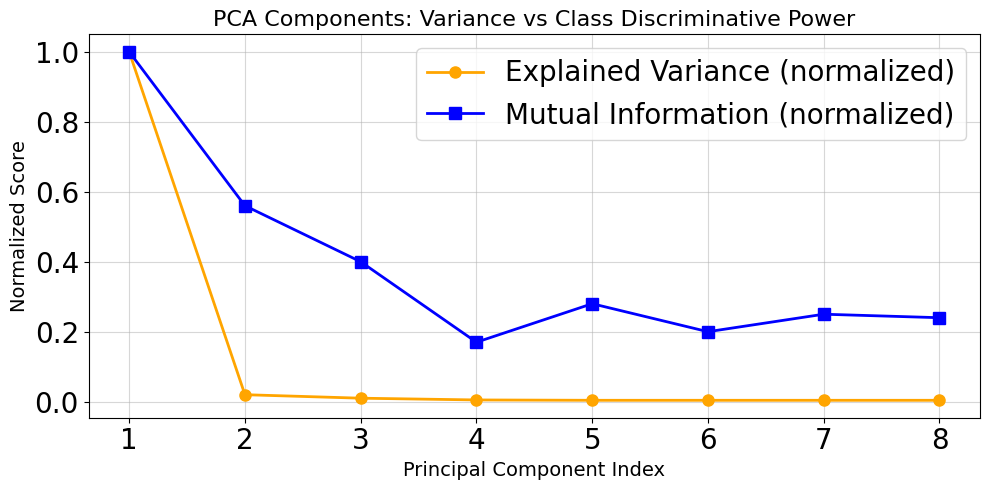}
    \caption*{(a) PCA - 8 components}
  \end{minipage}%
  \hfill
  \begin{minipage}[t]{0.48\columnwidth}
    \centering
    \includegraphics[width=\linewidth]{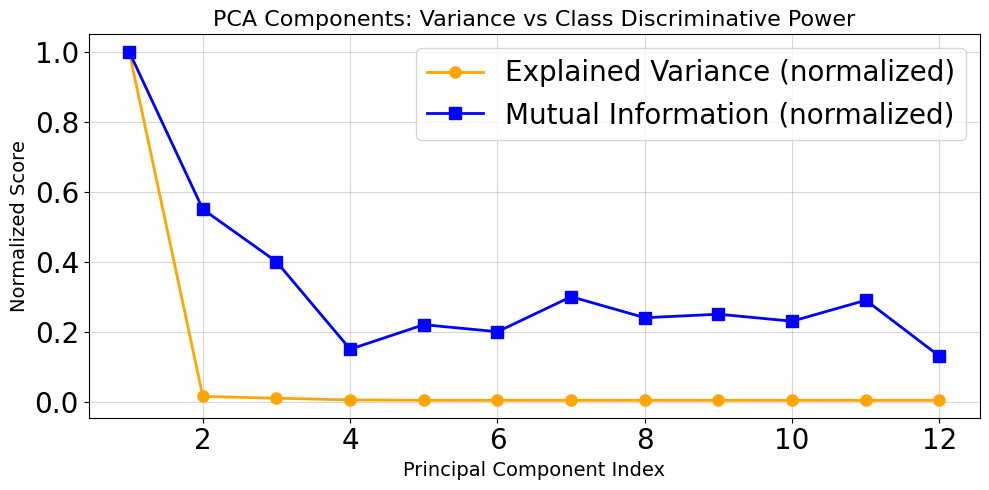}
    \caption*{(b) PCA - 12 components}
  \end{minipage}

  \vspace{0.4cm} 

  \begin{minipage}[t]{0.48\columnwidth}
    \centering
    \includegraphics[width=\linewidth]{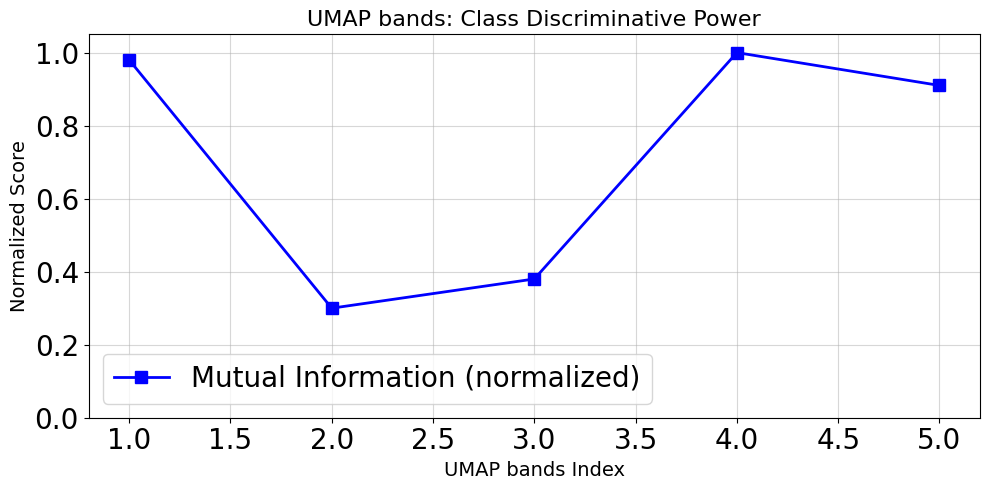}
    \caption*{(c) UMAP - 5 components}
  \end{minipage}%
  \hfill
  \begin{minipage}[t]{0.48\columnwidth}
    \centering
    \includegraphics[width=\linewidth]{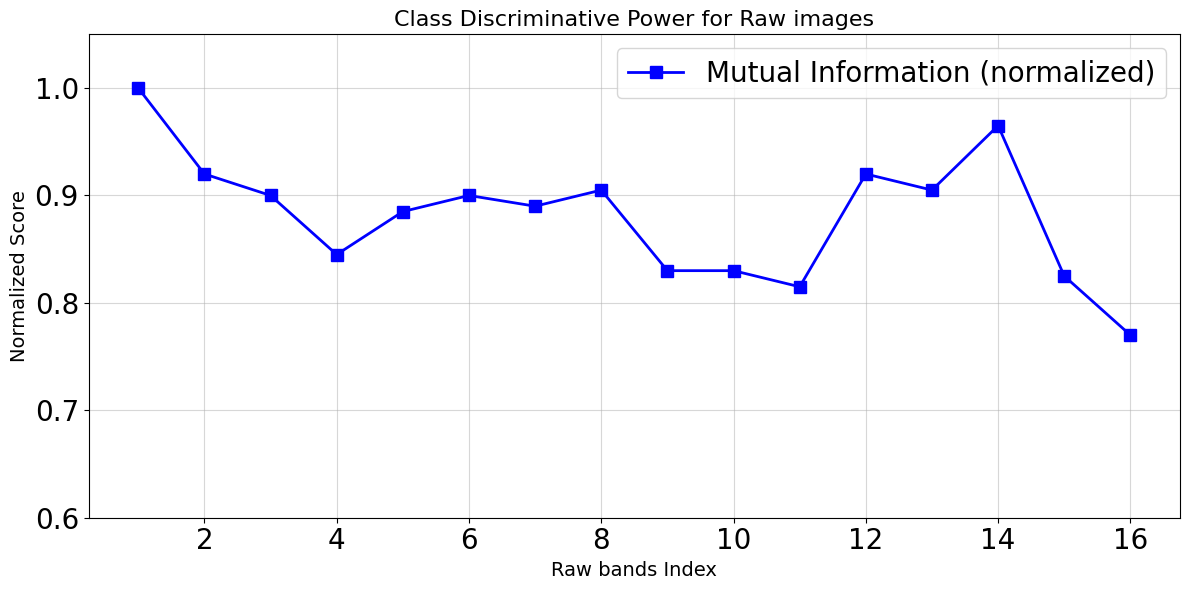}
    \caption*{(d) Raw data - 16 bands}
  \end{minipage}

  \caption{Comparison of PCA, UMAP, and raw hyperspectral data. The orange color is consistently used for mutual information across all plots.}
  \label{fig:comparison_plots}
\end{figure}

PCA, though popular for dimensionality reduction, may underperform in supervised tasks as it preserves variance without considering class separability. This can discard key spectral features and disrupt spatial-spectral patterns, reducing CNN performance and class-discriminative accuracy.

To better understand these limitations, we conducted a component-wise analysis of the PCA output using two complementary criteria: the explained variance ratio and the Mutual Information (MI) score~\cite{vergara2014review}. The explained variance ratio indicates the amount of spectral information retained by each component, while the MI score evaluates its relevance for classification by measuring the dependency between each PCA component and the class labels. This combined analysis highlights components that, despite having low variance, contain valuable discriminative information that would be overlooked by a purely variance-based selection.

Figures~\ref{fig:comparison_plots}~(a) and~(b) show the PCA results with 8 and 12 components, where the x-axis represents component indices and the y-axis two normalized metrics: explained variance  and mutual information. Figures~\ref{fig:comparison_plots}~(c) and~(d) display the mutual information of the five UMAP components and the 16 raw spectral bands, respectively.

The raw data (Figure~\ref{fig:comparison_plots}~(d)) contains several highly informative bands, confirming its rich spectral structure. UMAP (Figure~\ref{fig:comparison_plots}~(c)) effectively condenses this information into fewer components, preserving class-relevant features through its non-linear mapping. In contrast, PCA prioritizes variance, and some low-variance components exhibit moderate mutual information, showing that variance alone does not guarantee class relevance.

Overall, UMAP and raw data provide stronger and more consistent class-discriminative information, while PCA requires careful component selection for supervised tasks.

\subsection{Stones dataset}

\begin{table}[h!]
\centering
\resizebox{\linewidth}{!}{%
\begin{tabular}{|l|c|c|c|c|c|c|c|}
\hline
\textbf{Method} & \textbf{Bands} & \textbf{Mean SH} & \textbf{Err SH} & \textbf{Mean S$_A$} & \textbf{Err S$_A$} & \textbf{Mean S$_B$} & \textbf{Err S$_B$} \\
\hline
RAW   & 64 & 0.598 & 0.008 & 0.528 & 0.006 & 0.964 & 0.003 \\ 
NMF   & 20 & 0.596 & 0.006 & 0.512 & 0.011 & 0.963 & 0.004 \\ 
PCA   & 20 & 0.617 & 0.005 & 0.538 & 0.006 & 0.964 & 0.003 \\ 
UMAP  & 10 & 0.641 & 0.007 & 0.559 & 0.008 & 0.972 & 0.002 \\ 
UMAP  & 20 & \textbf{0.677} & 0.003 & \textbf{0.614} & 0.008 & \textbf{0.974} & 0.002 \\ 
UMAP  & 30 & 0.636 & 0.006 & 0.552 & 0.005 & 0.970 & 0.001 \\ 
\hline
\end{tabular}%
}
\caption{Performance metrics (mean and standard error) for different dimensionality reduction methods and band settings.}
\label{tab:score_summary}
\end{table}


Table \ref{tab:score_summary} shows that our UMAP-based dimensionality reduction consistently outperforms both traditional methods (PCA, NMF) and raw hyperspectral data. Using only 20 UMAP-derived bands yields the best results across all metrics (Mean SH = 0.677, Mean A = 0.614), surpassing the raw data baseline (0.598 and 0.528, respectively). UMAP with 10 bands underperforms due to information loss, while 30 bands slightly decrease performance, suggesting the reintroduction of noise. PCA and NMF were tested with the same number of bands (20) for a fair comparison. However, both methods fall behind UMAP, especially in Mean SH and Mean A, confirming that linear projections are insufficient for complex hyperspectral XRT data.

\subsection{Indian pines dataset} 

\begin{table}[H]
\centering
\begin{tabular}{|l|c|}
\hline
\textbf{Method} & \textbf{Accuracy (\%)} \\
\hline
\textbf{UMAP-CNN  }          & \textbf{98.80}  \\
T-SNE-CNN\cite{8876672}        & 97.89 \\
PCA-CNN             & 96.79 \\
DC-CNN \cite{zhang2017spectral} & 95.50 \\
DR-CNN \cite{8876672}           & 94.93 \\
Multiscale-CNN      & 87.42 \\
1D-CNN \cite{chen2016deep} & 66.36 \\
SVM \cite{mercier2003support}    & 74.81 \\
RF-200 \cite{joelsson2005random}  & 61.89 \\
\hline
\end{tabular}
\caption{Indian Pines classification results.}
\label{tab:classification_results}
\end{table}

Table~\ref{tab:classification_results} shows that UMAP-CNN achieves the best accuracy, surpassing both t-SNE-CNN and PCA-CNN. By capturing non-linear spectral structures that PCA cannot, UMAP provides more discriminative features with the same CNN. It is also more scalable and efficient than t-SNE, which becomes computationally prohibitive on larger datasets.


\section{Conclusion}
UMAP-based approaches consistently outperform traditional linear methods like PCA and NMF in our experiments. This is expected for X-ray transmission spectroscopy, where energy bands exhibit strong non-linear behavior, far more than in optical HSI. By approximating dataset topology through graph construction, our method captures these non-linear correlations, improving model performance. This work serves as a proof of concept, showing that accounting for non-linearities is essential in processing this type of data and motivating further research in topological data analysis for more accurate and robust representations.

\bibliographystyle{IEEEbib}
\bibliography{bib}

\end{document}